\documentclass[letterpaper, 10 pt, conference]{ieeeconf}  % Comment this line out
\usepackage{graphicx} 
\usepackage{lscape, afterpage, threeparttable}

\IEEEoverridecommandlockouts                              % This command is only
\title{\LARGE \bf
A Systematic Comparison of Deep Learning Architectures in an Autonomous Vehicle
}

\author{Michael Teti$^{1 \dagger}$, William Edward Hahn$^{1}$, Shawn Martin$^{2}$, Christopher Teti$^{3}$, and Elan Barenholtz$^{1}$
\thanks{$^{1}$Center for Complex Systems and Brain Sciences, Florida Atlantic \indent \indent University, 777 Glades Road, Boca Raton, FL 33431, USA}
\thanks{$^{2}$College of Computer and Information Science, Northeastern University, \indent \indent 360 Huntington Ave, Boston, MA 02115, USA}
\thanks{$^{3}$Department of Ocean and Mechanical Engineering, Florida Atlantic \indent \indent University, 777 Glades Road, Boca Raton, FL 33431, USA}
\thanks{$^{\dagger}$ {\tt\small mteti@fau.edu}}} % elan.barenholtz@fau.edu}}} % for anonymity

\begin{document}

\maketitle
\thispagestyle{empty}
\pagestyle{empty}

%%%%%%%%%%%%%%%%%%%%%%%%%%%%%%%%%%%%%%%%%%%%%%%%%%%%%%%%%%%%%%%%%%%%%%%%%%%%%%%%
\begin{abstract}
Self-driving technology is advancing rapidly --- albeit with significant challenges and limitations. This progress is largely due to recent developments in deep learning algorithms. To date, however, there has been no systematic comparison of how different deep learning architectures perform at such tasks, or an attempt to determine a correlation between classification performance and performance in an actual vehicle, a potentially critical factor in developing self-driving systems. Here, we introduce the first controlled comparison of multiple deep-learning architectures in an end-to-end autonomous driving task across multiple testing conditions. We used a simple and affordable platform consisting of an off-the-shelf, remotely operated vehicle, a GPU-equipped computer, and an indoor foam-rubber racetrack. We compared performance, under identical driving conditions, across seven architectures including a fully-connected network, a simple 2 layer CNN, AlexNet, VGG-16, Inception-V3, ResNet, and an LSTM by assessing the number of laps each model was able to successfully complete without crashing while traversing an indoor racetrack. We compared performance across models when the conditions exactly matched those in training as well as when the local environment and track were configured differently and objects that were not included in the training dataset were placed on the track in various positions. In addition, we considered performance using several different data types for training and testing including single grayscale and color frames, and multiple grayscale frames stacked together in sequence. With the exception of a fully-connected network, all models performed reasonably well (around or above 80\%) and most very well ($\sim$95\%) on at least one input type but with considerable variation across models and inputs. Overall, AlexNet, operating on single color frames as input, achieved the best level of performance (100\% success rate in phase one and 55\% in phase two) while VGG-16 performed well most consistently across image types. Performance with obstacles on the track and conditions that were different than those in training was much more variable than without objects and under conditions similar to those in the training set. Analysis of the model's driving paths found greater consistency within vs. between models. Path similarity between models did not correlate strongly with success similarity. Our novel pixel-flipping method allowed us to create a heatmap for each given image to observe what features of the image were weighted most heavily by the network when making its decision. Finally, we found that the variability across models in the driving task was not fully predicted by validation performance, indicating the presence of a `deployment gap' between model training and performance in a simple, real-world task. Overall, these results demonstrate the need for increased field research in self-driving. \\

\textit{Keywords} --- End-to-End Control Systems, Computer Vision, Machine Learning, Tensorflow, Autonomous Vehicles

\end{abstract}

%%%%%%%%%%%%%%%%%%%%%%%%%%%%%%%%%%%%%%%%%%%%%%%%%%%%%%%%%%%%%%%%%%%%%%%%%%%%%%%%%%%%%%%%
\section{BACKGROUND}

 Self-driving technology is advancing rapidly --- albeit with significant challenges and limitations. This progress is due in large part to the success of GPU-driven deep learning algorithms. However, many competing architectures and techniques for deep learning are currently available and/or in development, and little scientific research has been published that explicitly assesses best practices and outcomes across different learning approaches and models. This is due, in part, to the lack of self-driving data and results available to the academic research and larger public communities. The cost of such real-world systems (e.g. road-ready full size vehicles outfitted with broad arrays of sensors) can be prohibitive, making them off-limits to everyone but a few select large research institutions and, more commonly, private commercial ventures, who may not be encouraged to share their results with the broader research community because of commercialization concerns. Smaller systems (i.e. systems that are not the size of an actual automobile), on the other hand, may suffer from hardware limitations for onboard computers that are not capable of running state-of-the-art deep learning models. 

One of the most promising approaches to full autonomous performance is the use of so-called `end-to-end' learning models that are trained on sensor inputs, paired with human behavioral outputs, without the need for explicitly encoding intermediate representations. To date, only a few publications of which we are aware have used a deep neural network in an end-to-end fashion to control a real autonomous vehicle \cite{DBLP:journals/corr/BojarskiTDFFGJM16} \cite{bojarski2017explaining} \cite{yang2018end} \cite{xu2017end} \cite{sotoyolo}. However, these studies provide limited information regarding the model's training and performance on the road tests. Bojarski et al. \cite{DBLP:journals/corr/BojarskiTDFFGJM16} only report the results of a single trip on a real road without any description of basic features such as the miles traveled, nature of the roadway, conditions, training time, etc. Similarly, Yang et al. \cite{yang2018end} provide no quantitative results or a sufficient description of of the road test and tasks performed at all, and Xu et al. \cite{xu2017end} and Soto et al. \cite{sotoyolo} do not test their model in a vehicle at all. Furthermore, none of these provide any comparison across different model types and/or training protocols. Thus, these studies are of limited utility in establishing best practices for development. 

Other published results on autonomous driving are not focused on end-to-end supervised learning and/or not tested in actual vehicles but are instead concerned with specific sub-problems of self-driving, such as navigating in sub-optimal weather conditions \cite{Lee2018}, pedestrian detection \cite{mujahed2018admissible} \cite{zhang2017citypersons}, traffic light/obstacle detection \cite{hane2015obstacle} \cite{ramos2017detecting}, mind wandering detection \cite{baldwin2017detecting}, and the classification and/or segmentation of traffic scenes \cite{agarwal2017real} \cite{kundu2016feature}. These studies are generally performed on public datasets taken from dashcam videos such as the Udacity self-driving datasets \cite{ud} \cite{ud2}, the KITTI dataset \cite{Fritsch2013ITSC}, the more recently released SAIC \cite{yang2018end}, CityScapes \cite{cordts2016cityscapes}, BDD100k \cite{berkeley_selfdriving_dataset}, and Apollo \cite{apollo} datasets, or video games \cite{martinez2017beyond}. These approaches may lead to a potential ``deployment gap'' between a model's performance during training and validation --- what is essentially a traditional image classification task --- and its behavior in an embedded control system operating in the real world. In particular, once deployed, a self-driving model's behavior will also determine its \textit{inputs} which may end up being poorly represented by the human-generated dataset used to train and validate the models. As a result, there has been a recent effort by some industry leaders and the United States Department of Transportation to create a rigorous protocol for testing a self-driving technology's competency in an actual automobile, as there has already been one incident in which a self-driving vehicle struck and killed a pedestrian in Arizona \cite{uberaccident}. One such testing protocol consists of a ``91-acre, closed course testing facility . . . set up like a mock city'' that includes everything from highways to suburban driveways and railroad crossings \cite{cerf2018comprehensive}. Of course, access to such resources is highly limited and to date no systematic studies have been reported comparing different model performance in deployed driving tasks. 

Here we introduce the first (to our knowledge) systematic, real-world comparison of autonomous driving performance across multiple, contemporary deep neural networks and training data types. We use a simple, easily replicated platform, assembled from commercially available components (all hardware and software specs are described below; software is publicly available as a Docker repository). The setup consists of an off-the-shelf, remotely operated vehicle (Brookstone Rover 2.0), a GPU equipped computer and an indoor foam-rubber racetrack. The vehicle communicates with the computer over wifi in order to send its camera images to the computer. The images are then run through a trained neural network in real time in order to output an action decision that the computer sends back to the vehicle over the wifi network. This setup allows us to test computing intensive deep learning models without the need to ``onboard'' the GPU hardware. We used this platform to train seven different neural networks, across three image input classes, on data from multiple humans driving around the track. We then compared autonomous performance on the track under identical experimental conditions for each of the 21 (7 architectures $\times$ 3 image input classes) conditions. We report performance along multiple metrics including the percentage of successful loops (i.e. without crashing) and the average time in seconds needed to complete a loop. 

%%%%%%%%%%%%%%%%%%%%%%%%%%%%%%%%%%%%%%%%%%%%%%%%%%%%%%%%%%%%%%%%%%%%%%%%%%%%%%%%%%%%%%%%

\section{METHODS}

We compared the driving performance of multiple network architectures, which were chosen to reflect the diverse types of architectures employed in recent years as well as some older ones, in driving a remote controlled vehicle around a track after being trained on human driving data. The tested architectures included: 1) a three hidden layer fully-connected network, 2) a simple convolutional neural network (CNN) with two convolutional/pooling layers followed by two fully-connected layers, 3) AlexNet, \cite{NIPS2012_4824}, 4) a slightly modified version of VGG-16 \cite{DBLP:journals/corr/SimonyanZ14a}, 5) Inception-V3 \cite{DBLP:journals/corr/SzegedyLJSRAEVR14}, 6) a version of the ResNet architecture which we refer to as ResNet-26 \cite{DBLP:journals/corr/HeZRS15}, and 7) a Long Short-Term Memory (LSTM) network \cite{hochreiter1997lstm}. The details of each network are described below in detail. Each network's architecture, as well as the training procedure can be viewed at {\tt\small https://github.com/mpcrlab/AIRover}. 

Another goal of the current study was to determine what kinds of information are most useful in end-to-end training of an autonomous driving system. For example, how helpful is it to include color information, which involves three times as much input information as grayscale? To assess this, we included three input types used as training and test data for the different models: 1) single grayscale video frame, 2) single color video frame, and 3) the current grayscale video frame plus past grayscale video frames concatenated along the channel dimension (which we term `framestack') as input to each different network. The framestack method provides a simple method for incorporating temporal information without the need for an architecture that is specifically designed to incorporate sequential information (e.g. CNNs). These three input classes were chosen to determine whether spatial, color, or temporal information is more useful for such tasks, a consideration when designing low-power, smaller systems that may not be able to afford to utilize all three feature modalities. Note that the framestack approach that we use here provides a method for including temporal sequence information in a simpler manner than typical approaches, such as recurrent neural networks. This allowed us to test the role of temporal information, using the same architectures as we used for the individual images. In addition, it introduces a novel, potentially simpler approach to incorporating temporal information in self-driving applications.

\subsection{Experimental Setup}
\begin{figure}[ht]
\begin{centering}
\includegraphics[width=\linewidth]{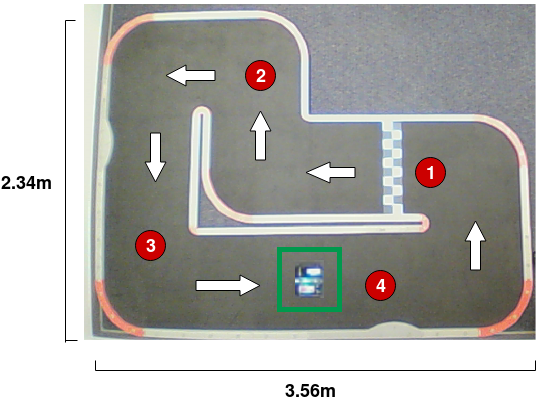}
\caption{The track used to train and test each network. The vehicle was trained and tested on its ability to navigate the track successfully in the direction indicated by the white arrows. The four test positions are indicated by the red circles, and the vehicle is contained within the green box.}
\label{fig:track}
\end{centering}
\end{figure}

\begin{figure}[ht]
\begin{center}
\includegraphics[width=5cm]{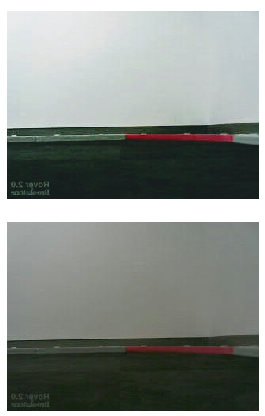}
\caption{A sample frame from the vehicle's camera taken approximately from its position in Fig. 1. The top frame was taken under the high-light condition, and the bottom was taken under the low-light condition. As can be seen in these images, the vehicle's camera was very narrow and could only capture the scene directly in front of it, which added difficulty to the task.}
\label{fig:frame}
\end{center}
\end{figure}

To test each network architecture in a self-driving task, we used a 3.56m $\times$ 2.34m L-shaped foam racetrack \cite{rcptracks} and a Brookstone Rover 2.0 \cite{brookstone} (Fig. \ref{fig:track}). Each $240 \times 320$ color video frame (Fig. \ref{fig:frame}) collected by the vehicle's single, built-in camera (which was set to collect 30 frames per second) was sent over wifi to a computer containing two GeForce GTX 1080 TI GPUs. 

\subsection{Training Protocol}
To create a supervised dataset on which to train each network, multiple humans drove the vehicle a single direction around the track (Fig. \ref{fig:track}) under variable lighting conditions (during one half all of the overhead room lights were on, and during the other half only one-third of the room's lights were on). We recorded each video frame along with the action --- left pivot, right pivot, forward, or backward --- the human performed at that frame. The dataset, which totaled approximately 250,000 frames and their respective labels, was composed of a validation dataset of $\sim7,000$ frames, which was taken from a completely separate test run than those used in training, and a training dataset of $\sim$243,000 frames. The validation set was used to test the network every 100 training iterations.

Each network was tested and trained on these same validation and training sets for 6,000 training iterations. The number of training iterations was chosen such that slower-learning networks would have sufficient training time to learn while still controlling for the number of training iterations, as many networks converged well before this point but did not overfit. Each training iteration began with a random video frame and the subsequent 79 video frames (80 in total) from the training set. The training batch size was determined by finding the maximum number of examples that the most resource-intensive network could handle on our GPU hardware (essentially double the chosen training examples due to one of our augmentation methods) and using that number of examples to to train each network.

Once the batch was randomly chosen from the training set, each frame was cropped by removing the top 110 rows (making the images $130 \times 320 \times 3$), and further operations were performed depending on the image processing method being employed. These operations are described below: 

\subsubsection{Grayscale}

For the video frames in the grayscale method, each frame was made grayscale and instance normalization \cite{DBLP:journals/corr/UlyanovVL17} was subsequently employed on each individual frame.

\subsubsection{Color}

For the color class, instance normalization \cite{DBLP:journals/corr/UlyanovVL17} was used on each color frame. 

\subsubsection{Grayscale Framestack}

Each video frame in this method began with the same operations as in the grayscale class. Each frame$_{t}$ in the batch was then paired with the frame$_{t-5}$ and frame$_{t-15}$ along the channel dimension. The human action at frame$_{t}$ was used as the label for the framestack training example. The intervals were chosen empirically by trying many different values and observing how well the trained vehicle navigated the track. These intervals are likely dependent, at least to some extent, on the frame rate of the camera(s), as well as the top speed of the vehicle. There is some existing research in which temporal correlations in video data were exploited in a similar manner \cite{pan2016learning}. 

After performing the appropriate operations, each image's height and width were zero-padded with 30 pixels and randomly cropped as in \cite{NIPS2012_4824}. After cropping, a copy of each image was created (essentially doubling the batch size), and white noise was added to each of the copies\footnote{We also tried to augment each data batch by flipping each image with respect to the vertical axis and changing the label accordingly, but this caused the vehicle's movement to be less continuous and its accuracy worse.}. The peak signal-to-noise ratio (PSNR) was computed over 500 random frames and their noise-augmented counterparts, and the average for each frame was 10.0dB. The batch was then sent to the neural network to continue the training iteration. Each network's weights were optimized using Tensorflow's Adam optimizer and a learning rate of 3e-5. 

\section{Architectures Tested} 
We tested seven different models, described below. (A description of each model's layer architecture is included in Table 2). For the CNNs besides ResNet-26, these were initialized with random values from a uniform distribution without scaling variance as described in \cite{DBLP:journals/corr/Sussillo14}, and weights in all fully-connected layers were initialized with random values taken from a truncated normal distribution as in \cite{gers1999learning}. The weights in the convolutional filters in ResNet-26 were initialized with random values such that the variance of the inputs would be constant as in all other networks' convolution layers, but, instead of taking these values from a uniform distribution, they were taken from a truncated normal distribution as in \cite{DBLP:journals/corr/HeZR015}. A weight decay \cite{krogh1992simple} of 0.001 was employed in all convolution and fully-connected layers. All convolution layers utilized a rectified linear activation function \cite{dahl2013improving}, and all fully-connected layers employed a hyperbolic tangent activation function \cite{kalman1992tanh}, except for those at the end of VGG-16 which used rectified linear activation functions. Those networks with max pooling \cite{wersing2003learning}, with the exception of the 2-layer CNN, use overlapping pooling \cite{NIPS2012_4824} with a kernel size of $3\times3$ and a stride of 2. Dropout \cite{srivastava2014dropout} of fully-connected nodes, which is used to reduce overfitting, was utilized in all networks except ResNet-26, with dropout probabilities of 0.5 and 0.0 for training and testing, respectively (Inception-V3 contained a dropout probability of 0.6 for training). Instead of using dropout to reduce overfitting of fully-connected layers, ResNet-26 employs global average pooling on the last layer of feature maps \cite{DBLP:journals/corr/LinCY13}, which reportedly helps the network's generalization ability. The output layer of every network consisted of four fully-connected nodes and a softmax activation function \cite{dunne1997pairing}.

\subsection{Fully-Connected Network}
Perhaps the most basic deep artificial neural network, the fully-connected network consists of the input layer, three hidden layers, and the output layer. The input layer contains 124,800 input nodes for color images ($130 \times 320 \times 3$). Each hidden layer contains 64 nodes, which are each connected to every node in the previous and subsequent layers, with a hyperbolic tangent activation function \cite{kalman1992tanh}, $\ell_2$ regularization \cite{DBLP:journals/corr/Laarhoven17b} to reduce overfitting and complexity, and weight decay of 0.001 \cite{krogh1992simple}. Dropout \cite{srivastava2014dropout} is also applied after each hidden layer to decrease the chance of overfitting the training data. The first fully-connected network to be employed successfully in an autonomous vehicle was developed by Dean Pomerleau in 1989 as part of the ALVINN (Autonomous Land Vehicle in a Neural Network) project \cite{pomerleau1989alvinn}.

\subsection{2-layer CNN}
This architecture was chosen because it is perhaps one of the simplest convolutional networks and would, as a result, allow for a close comparison between a fully-connected architecture and a CNN architecture. $\ell_2$ regularization \cite{DBLP:journals/corr/Laarhoven17b} is used in both convolution layers. After each convolution layer, $2 \times 2$ max pooling \cite{riesenhuber1999hierarchical} \cite{wersing2003learning} with stride 2 and local response normalization \cite{NIPS2012_4824}, which encourages sparsity, were applied.

\subsection{AlexNet}
First published in 2012, AlexNet \cite{NIPS2012_4824} remains one of the most well-known and widely used deep neural networks to date, which is greatly due to its remarkable performance on the ImageNet Large Scale Visual Recognition Challenge in 2010 \cite{ILSVRC15}. Since its publication, AlexNet has been used on many tasks, including object detection \cite{DBLP:journals/corr/GirshickDDM13}, image segmentation \cite{DBLP:journals/corr/LongSD14}, and video classification \cite{KarpathyCVPR14}, to name a few. Following these applications and achievements of AlexNet, the computer vision and neural network communities were spurred to move from the engineering of features to the engineering of networks \cite{DBLP:journals/corr/XieGDTH16} and create deeper, more elaborate networks that could perform even better at such tasks. 

\subsection{VGG-16}
Larger, more elaborate networks, however, often pose additional challenges due in part to the increased number of hyper-parameters, which must still be chosen relatively carefully at this time. For example, the stride size, filter size, and number of filters in a convolution layer have an effect on the performance of the network. The VGG architecture \cite{DBLP:journals/corr/SimonyanZ14a} attempts to address the issue of choosing different stride and filter sizes by using a stride of one and a filter size of $3\times3$ for all convolution layers. Thus, this style of architecture reduces the number of hyper-parameters --- despite its greater depth --- than its predecessor AlexNet by stacking ``building blocks of the same shape . . . which increases simplicity and reduces the chance of overadapting the architecture to a specific dataset'' \cite{DBLP:journals/corr/XieGDTH16}. The ability of VGG-nets to generalize to different tasks has been shown in many applications \cite{DBLP:journals/corr/DonahueJVHZTD13} \cite{DBLP:journals/corr/Girshick15} \cite{Pinheiro:2015:LSO:2969442.2969462} \cite{DBLP:journals/corr/XiongDHSSSYZ16}. 

\subsection{Inception-V3}
In contrast to the VGG-style architectures, the Inception architecture \cite{DBLP:journals/corr/SzegedyLJSRAEVR14} \cite{DBLP:journals/corr/SzegedyIV16} \cite{DBLP:journals/corr/SzegedyVISW15} contains hand-crafted topologies with many varying hyper-parameters while still exhibiting low model complexity \cite{DBLP:journals/corr/XieGDTH16} and high performance on an array of tasks \cite{esteva2017dermatologist} \cite{7952132} \cite{gkioxari2016chained}. These architectures, including Inception-V3, which we use here, all operate on the principle of splitting the feature map outputs of certain layers into multiple different streams of operations (represented by the dashed lines in Table 2) and subsequently merging their outputs together via concatenation.

\subsection{ResNet-26}
The ResNet architecture \cite{DBLP:journals/corr/HeZRS15} builds off of the splitting/merging strategy of the Inception architectures and the simple, block-template style of VGG nets. These networks are composed of ``residual blocks'', where a template of convolutions is repeated a set number of times, and after each repetition, the features that served as input to that specific repetition are added to the output of the repetition. This is possible because of the block-template structure employed in VGG nets, as the output of a residual block often contains the same dimensions of the input of the block. When this is not the case (i.e. when the stride is greater than one or when increasing the number of filters) the input is downsampled via average pooling with a stride length of two and/or a linear transformation is used to increase the channel dimension of the input \cite{DBLP:journals/corr/HeZRS15} respectively. After every convolutional layer, batch normalization \cite{DBLP:journals/corr/IoffeS15} and a rectified linear activation function \cite{dahl2013improving} are applied to the output (except when adding the identity of the previous block, in which the activation function comes after the addition). The model complexity, measured in FLOPs and number of parameters of ResNets is extremely low relative to other CNNs (Table 2), yet these networks are still able to perform well on and generalize to different tasks \cite{DBLP:journals/corr/XiongDHSSSYZ16} \cite{akbar2018determining} \cite{lu2018deep} \cite{rezende2017malicious} \cite{jung2017resnet} \cite{zhang2017combination}. 

\subsection{LSTM}
Since their development in the mid-1990s in the context of language and writing processing, LSTMs have proven to be well suited to an array of problems that contain sequential data, as they are able to capture both long- and short-term dependencies in such data. They also are less susceptible to the problems encountered by simple recurrent networks \cite{hochreiter2001gradient}. As a result, they have been used in applications from handwriting classification \cite{doetsch2014fast} \cite{graves2009novel} to handwriting generation \cite{graves2013generating} and speech translation \cite{luong2014addressing}. Each node in these networks has four gates: input, output, and input modulation which use the sigmoid activation function as in \cite{karlik2011performance}, and the forget gate which uses the hyperbolic tangent activation function \cite{kalman1992tanh}. These gates work in conjunction with each other to help regulate which information enters the cell state, which is able to hold long-term dependencies, and a hidden state, which captures the short-term dependencies. The typical input for such networks is an $m \times n$ matrix, where each row is the next timestep in the data and each column is a dimension in those timesteps. Here, we treat each image as this matrix, as though the rows of the image are the timesteps with $320 \times 3$ dimensions for color and grayscale framestack images and 320 dimensions for grayscale images. We do this by concatenating the input channels along the column dimension\footnote{It is worth noting that we initially attempted concatenation along the row dimension with poor results.}. The network we use here has two hidden LSTM layers each containing 500 nodes, where each layer is essentially comprised of four fully-connected layers representing each gate. Each node in the first of these hidden layers outputs a sequence of hidden states corresponding to the number of time-steps (image rows), whereas each node in the second hidden layer returns one output for the whole sequence it was given.

\section{Testing Protocol}
After each network completed training, it was then used to control the vehicle autonomously at a constant speed around the track. To measure the performance of each network, the vehicle was placed at four different positions around the track (Fig. \ref{fig:track}) and driven autonomously for 10 trials at each position, totaling 40 test trials per network. Five of these ten trials at each position were performed under a high-light condition (all lights on; Fig. \ref{fig:frame}, top) and five with a low-light condition (one-third of the lights on, Fig. \ref{fig:frame}, bottom).

Every time the testing position was changed (every 10 trials), the vehicle's batteries, which were new and unused at the start of this research, were replaced with fully-charged ones. A camera, which was fixed to the ceiling and faced down toward the track, was used to film each test run. Along with this video, the time of each trial was taken, and it was recorded whether the vehicle successfully completed a single lap or not. A trial was ended when one of four circumstances occurred: 1) the vehicle completed a lap and made it back to its starting position, 2) the vehicle turned around and went the wrong direction three times in the same trial (most models eventually turned back around and righted themselves when this occurred), 3) the vehicle hit the wall and/or became stuck, or 4) the vehicle became stuck in an oscillatory back-and-forth motion without making net progress on the track for 10 consecutive seconds.

\begin{figure}[ht]
\begin{centering}
\includegraphics[width=\linewidth]{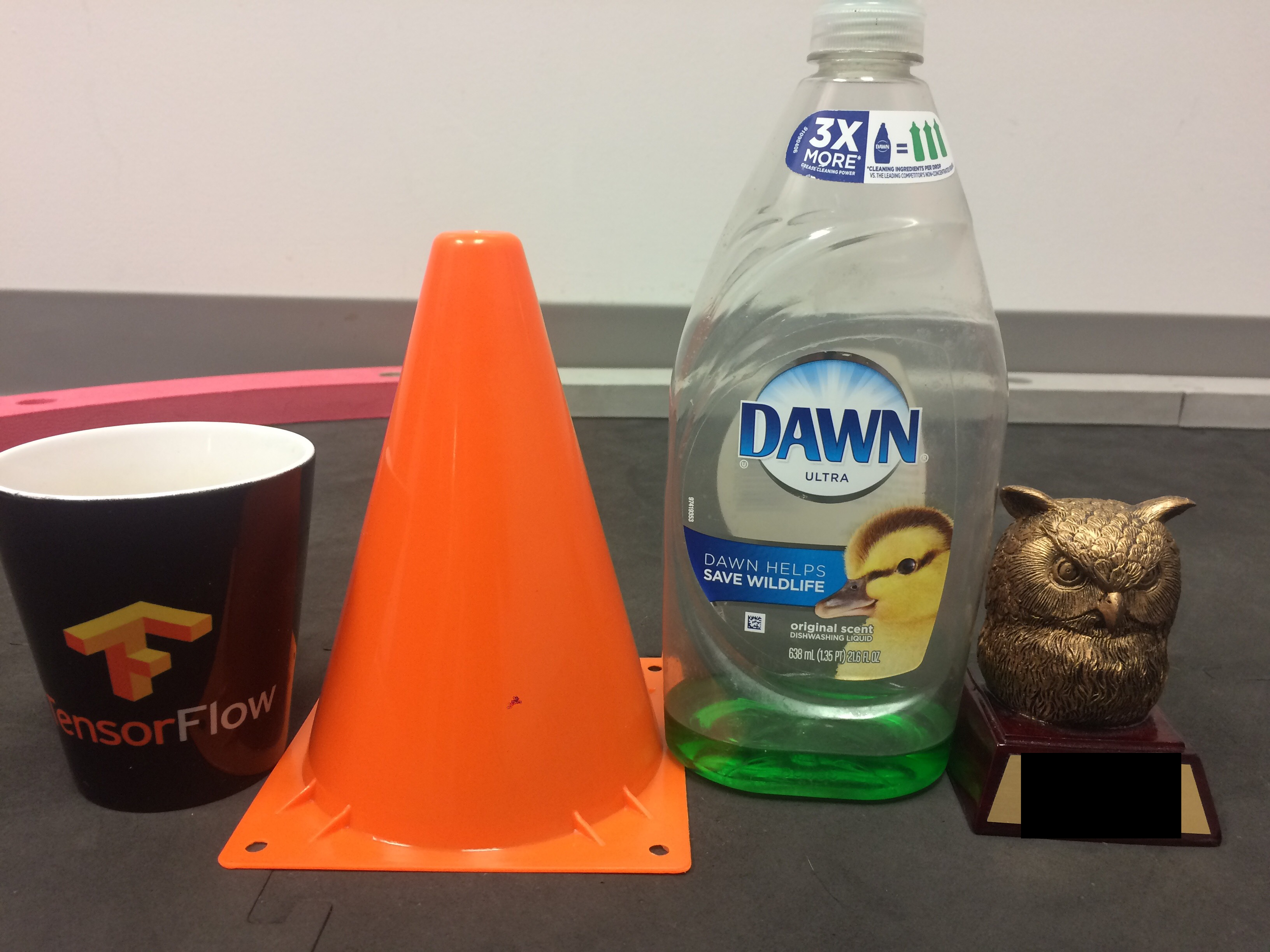}
\caption{The four objects used in the second testing trial of this research.}
\label{fig:obj}
\end{centering}
\end{figure}

Using the protocol above, each network was tested in two separate test trials. During the first testing trial, the same track shape and environment (i.e. room layout) that were used in the training data were also used in the testing phase. In the second testing trial, each network was tested under the same protocol but only using the input image method that enabled it to obtain the best performance in the first testing trial. Furthermore, during this second testing phase, the objects in the room (which were in the vehicle's field-of-view while it was driving) were rearranged, the track was configured in an oval shape instead of the L-shape, four diverse objects (pictured in Fig. \ref{fig:obj}) which were not present in the training data were placed randomly on the track, and a different vehicle of the same make and model of the training vehicle was used. For each of these test trials, a random number generator was used to determine 1) the number of objects placed on the track, 2) how far along the lap each object should be placed, 3) where each object was positioned relative to the middle of the path, and 4) how much the object was rotated. All of these parameters regarding object placement were consistent across all networks tested.

%To compare each network's performance to human performance in driving the vehicle, ten human subjects --- which ranged in age from 15-45 but were mostly college-aged --- were tested on their ability to 1) label individual frames according to the labels used for training the networks, and 2) drive the vehicle around the track from each of the four starting positions on the oval testing track while using only the video feed from the vehicle's camera. Of the ten subjects, three were authors of this research and were very familiar with the vehicle's driving, four had limited experience driving the vehicle previously, and the final three had no prior experience with the vehicle at all. 

\subsection{Performance Analysis}
The equation,

\begin{equation}
\textnormal{success rate} = \frac{\textnormal{\# of trials with lap completed}}{40}
\end{equation}
\\
was used as the primary metric to determine how each network performed on this task. 
% Using the success rate, we then calculate each network's "efficiency" by the following equation:

% \begin{equation}
% \textnormal{efficiency} = \frac{\textnormal{success rate}}{\textnormal{\# of parameters}}
% \end{equation}
% \\
% It is worth noting that the number of parameters and FLOPs in each model were highly correlated, so we only use parameters here.

%Because the vehicle's speed was constant for each network, the time it took to complete a successful lap was mainly a function of how continuous its motion was and how efficiently it navigated the track. For example, if the network caused the vehicle to move forward in a zig-zag fashion, it could still complete the lap successfully, but its mean time would be higher than if it continued in  a straight, continuous motion. Therefore, we calculate the mean time it took to complete a successful lap using the times recorded for each successful lap of a given network. 

The number of inferences per second each network could perform on images from this dataset was calculated by having the network perform 1000 inferences, obtaining the elapsed time and dividing by 1000. This procedure was performed five times for each network, and the average of these five trials is reported. This metric is potentially relevant to mobile/vehicular applications, where the ability of a network to perform a certain number of inferences per second is critical when moving at a high speed and/or difficult conditions.

\subsection{Path Analysis}
To further explore each network's effect on the vehicle's path, we used the videos taken by the ceiling camera and employed an object-tracking algorithm to determine the location of the center of the vehicle during every trial in testing phase one\footnote {We did not film the second testing phase.}. These coordinates were used to overlay colored dots on top of an image of the track to indicate where the vehicle had traveled over its 40 trials. Using these coordinates, it was possible to quantify path similarities and differences across runs between and within networks. To compare the paths taken in two different trials, the coordinates of the two trials were paired by time-step and by starting location. The average mean-squared distances between each of these corresponding points was then calculated for all of the pairs of points in the two compared trials. If one trial ended before the other (i.e. the vehicle failed to complete the lap due to a crash, etc.) the longer trial was shortened to match the length of the shorter trial. Using this method, we compared each trial to every other trial and obtained a single number representing the average path differences across the compared models.

\subsection{Hyper-Parameter Analysis}
In order to determine which hyper-parameter(s) were most important in determining a network's success rate in each testing phase, a random forest regression model \cite{breiman2001random} was trained on a number of meta variables surrounding each network with the goal of predicting the success rate in the respective testing phase. For each testing phase, we ran scikit-learn's \cite{scikit-learn} RandomForestRegressor 1000 times. Each time, the number of estimators and the maximum number of features to consider when looking for the best split were chosen randomly from the ranges [500, 5000] and [1, \textit{n}-1] respectively (where \textit{n} is the number of hyper-parameters input to the random forest model). All other parameters of the random forest were kept at the default values. After the forest is trained, it is possible to observe the importance of each feature in predicting the output, which allowed us to determine what hyper-parameters played a large role in determining success in each testing phase. We derived each feature's average rated importance across the 1000 runs. Some of the input variables considered by the random forest model were the number of FLOPs and parameters, the number of hidden layers in a network, the mean validation loss over the last 1200 training iterations, and the validation loss at the start of training. 

\subsection{Network ``Bias''}
One factor that is influential to generalization ability --- and a possible deployment gap --- is the ability of a network to avoid overfitting the training set. This is made more difficult when the frequencies of the labels (or actions) in the training set are very uneven (e.g. the forward action appeared much more than any other in the training set). In order to determine whether certain networks overfit to the distribution of actions in the training set, and how they were affected by this, the bias weights of the output layer were examined and compared against the actual distribution of labels in the dataset. This method helped determine whether a given network was `biased' toward a specific action based on the training set and how this `bias' affected its performance in both testing phases. 

\subsection{Spatial Distribution of Attention}
In order to gain further insight into the observed differences between tested models, we assessed which portions of the image each model deemed more important in order to make its behavioral decisions. To do so, we utilized a novel method loosely based on \cite{one-pixel} in which we systematically `flipped' the values of each pixel value in the input images, one by one, and observed the corresponding difference in the model outputs compared to the unaltered image. This serves to determine what pixels/regions of the image were more important in the network's classification. This method bears some similarity to other recently developed methods designed to make neural networks more interpretable --- often using methods such as deconvolutional layers \cite{zeiler2010deconvolutional} or layer-wise relevance propagation (LRP) \cite{bach2015pixel}. However, these other methods do not generalize well to all neural network architectures and their results are not always readily interpretable \cite{interpret_learning}. The method introduced here provides a simple and easily interpretable method for localizing the information in images across all models. The procedure consisted of five steps: 1) preprocess and perform inference on a given test image and record the output layer values and the action chosen, 2) loop through every pixel in the image and maximally flip its value (i.e. pixel values $\geq$ 128 were given the value 0 and those $<$ 128 were given the value 255), 3) perform preprocessing and inference on the image with the altered pixel, 4) calculate the MSE between the output layer values associated with the altered image and the unaltered image and record it along with the altered pixel's location, and 5) determine the action from the altered image's output and record whether changing that pixel caused the network to choose a different action. This procedure was performed over 50 images chosen randomly from a test dataset taken on the L-shaped track. To assess how easy it was to change the output action for a given network, we also calculate the average confidence value for the action chosen over 5000 images. Finally, we use the MSE for each pixel location in step 4 to create a heatmap in the image space that illustrated how important each pixel of the image was in changing the output. 

% To determine which network / image processing combination(s) performed best, we used the equation,

% \begin{equation}
% \textnormal{Performance} = \frac{\textnormal{Percent of successful laps}}{\textnormal{mean time of successful laps}} 
% \end{equation}
% \\
% where the fraction of successful laps is the number of laps successfully completed by a network divided by 40. We used this performance metric because the success rate of multiple networks was 100\%, so this alone was insufficient in determining which one performed better. A network's performance would decrease if that network took longer on average to complete a lap and/or successfully completed fewer laps, but it would increase if the network took longer to complete each lap and/or completed more laps successfully. As a result, the network that best optimized its speed/efficiency and accuracy would have the highest performance. Because the rover's speed was constant for each network, a network's mean speed was mainly a function of how continuous its motion was and how efficiently it navigated the track. For example, if the network caused the rover to move forward in a zig-zag fashion, it could still complete the lap successfully, but its mean time would be higher than if it continued in  a straight, continuous motion. 

%%%%%%%%%%%%%%%%%%%%%%%%%%%%%%%%%%%%%%%%%%%%%%%%%%%%%%%%%%%%%%%%%%%%%%%%%%%%%%%%%%%%%%%%%
\section{RESULTS}
\subsection{Validation Performance}
\begin{figure}
\begin{centering}
\includegraphics[width=\linewidth]{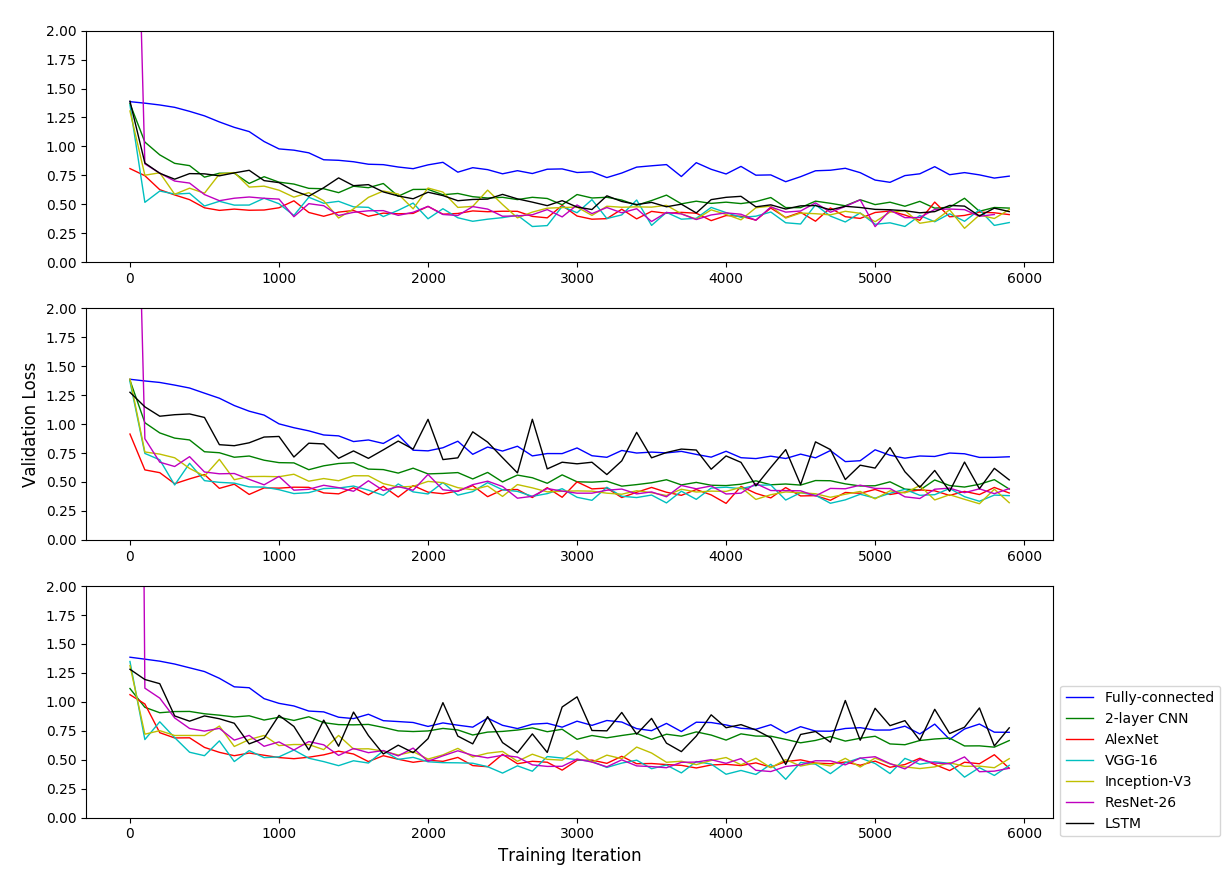}
\caption{Each network's validation loss over training with single grayscale frame (top), single color frame (middle), and grayscale framestack (bottom) as input.}
\label{fig:val_loss}
\end{centering}
\end{figure}

Fig. \ref{fig:val_loss} shows the validation loss of each model over the entire course of training for each of the three input types. For the single grayscale frame and single color frame conditions (top and middle), all of the models converged to moderate losses except for the fully-connected network, which yielded a loss that was approximately $2\times$ higher upon model convergence than all others. For the grayscale framestack input (bottom), the four contemporary CNNs show significantly reduced loss upon convergence compared with the three other models. 

\subsection{Success Rate}
\subsubsection{Testing Phase One}
Fig. \ref{fig:success_rates} (top) presents the success rate (i.e. the percentage of trials in which the lap was completed) across all of the tested architectures during testing phase one. Overall, the convolutional neural networks and LSTM vastly outperformed the fully-connected network. Within the contemporary CNNs, all of them achieved reasonably good success rates ($\sim$95\%) with at least one data type. However, only AlexNet, trained on single color video frames, achieved a perfect success rate over 40 trials. VGG-16 was found to be the most robust to the input class, as its success rate was the equally high for all three input image types.

Across the different data types, the color single frame yielded the best overall performance across models while the grayscale framestack lagged dramatically across most models. 

% Although success rate alone was very similar for many networks, ResNet-26 emerged as the best network when each network's efficiency (Fig. 6, top) was considered, followed by LSTM and Inception-V3.

\subsubsection{Testing Phase Two}
\begin{figure}
\begin{centering}
\includegraphics[width=\linewidth]{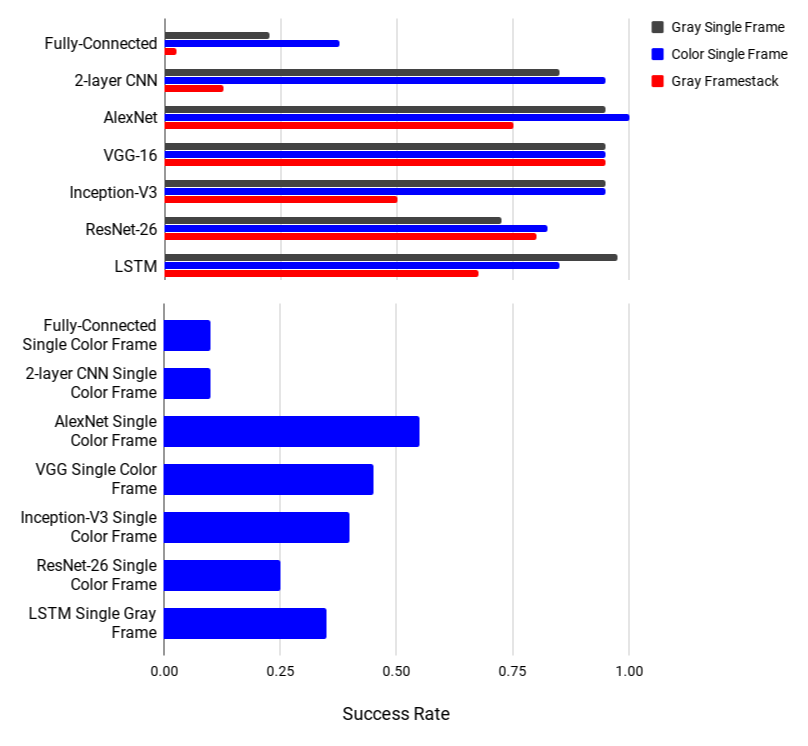}
\caption{The success rate of each network during test phase one (top) and test phase two (bottom).}
\label{fig:success_rates}
\end{centering}
\end{figure}

The success rates achieved by each network in testing phase two were much lower than those in phase one (Fig. \ref{fig:success_rates}, bottom). AlexNet exhibited the best success rate during this phase, completing 55\% of the 40 laps, followed by VGG-16 which completed 45\% of the 40 test laps (Fig. \ref{fig:success_rates}, bottom). ResNet-26 exhibited a disappointing performance during this phase, as it only completed 25\%, or 10, of its test laps. 

%When the human subjects used the vehicle's camera to drive around the track under the same protocol the networks were tested under, they achieved a higher success rate, both when the video feed was grayscale (success rate = 0.85) and color (success rate = 0.90).

\begin{figure}
\begin{centering}
\includegraphics[width=\linewidth]{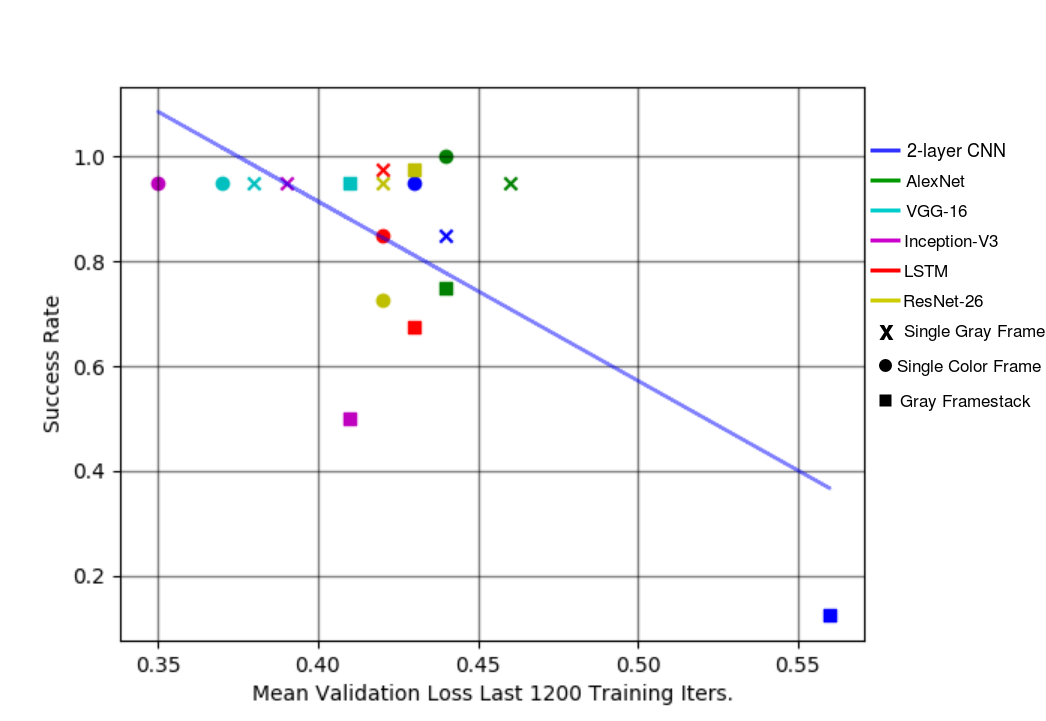}
\caption{The success rate as a function of the mean validation loss over training iterations 4800 through 6000 (fully-connected network not shown). R\textsuperscript{2} = 0.46.}
\label{fig:val_succ}
\end{centering}
\end{figure}

\begin{figure}
\begin{centering}
\includegraphics[width=\linewidth]{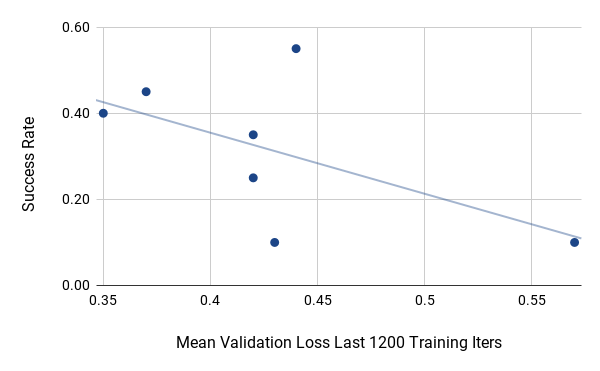}
\caption{Success rate of each network in testing phase two as a function of the respective mean validation loss over training iterations 4800 through 6000. R\textsuperscript{2} = 0.34.}
\label{fig:val_succ2}
\end{centering}
\end{figure}

In order to test for a possible 'deployment gap'---i.e., the extent to which offline training/validation does not predict real driving performance--- we calculated the mean validation loss for each model and input type for the last 1200 training iterations. Figures \ref{fig:val_succ} and \ref{fig:val_succ2} show the relationship between the validation loss and the respective success rate during testing phases one and two, respectively. As can be seen, many model/input types with similar validation losses (those between .4 and .45) demonstrate widely variable success rates in both testing phases, suggesting the presence of a deployment gap. For example, Inception-V3 trained on grayscale framestack input had the same validation loss as VGG-16 trained on the same input, but the former's success rate in testing phase one was 50\% and the latter's was 95\%. Furthermore, AlexNet trained on single color frames was the only network to achieve perfect success in testing phase one despite having one of the worst validation losses (Fig. \ref{fig:val_succ}), and an array of network/input combinations obtained 95\% success or better while their validation losses showed significant  variability (Inception-V3/color frame, VGG-16/color frame, VGG-16/grayscale frame, Inception-V3/grayscale frame, VGG-16/grayscale framestack, ResNet-26/grayscale frame, 2-layer CNN/color frame, and AlexNet/grayscale frame; Fig. \ref{fig:val_succ}). The most dramatic examples are AlexNet, using single grayscale frames and Inception-V3 using single color frames. These two models showed  validation losses that differed from each other by over 20\% but the same success rate in testing. Similar conclusions for testing phase two can be drawn in Fig. \ref{fig:val_succ2}, as many of the same models (i.e. 2-layer CNN, LSTM, ResNet-26, and AlexNet) performed very differently despite having a similar mean validation loss of $\sim$0.43.

\subsection{Path Analysis}
\begin{figure}
\begin{centering}
\includegraphics[width=\linewidth]{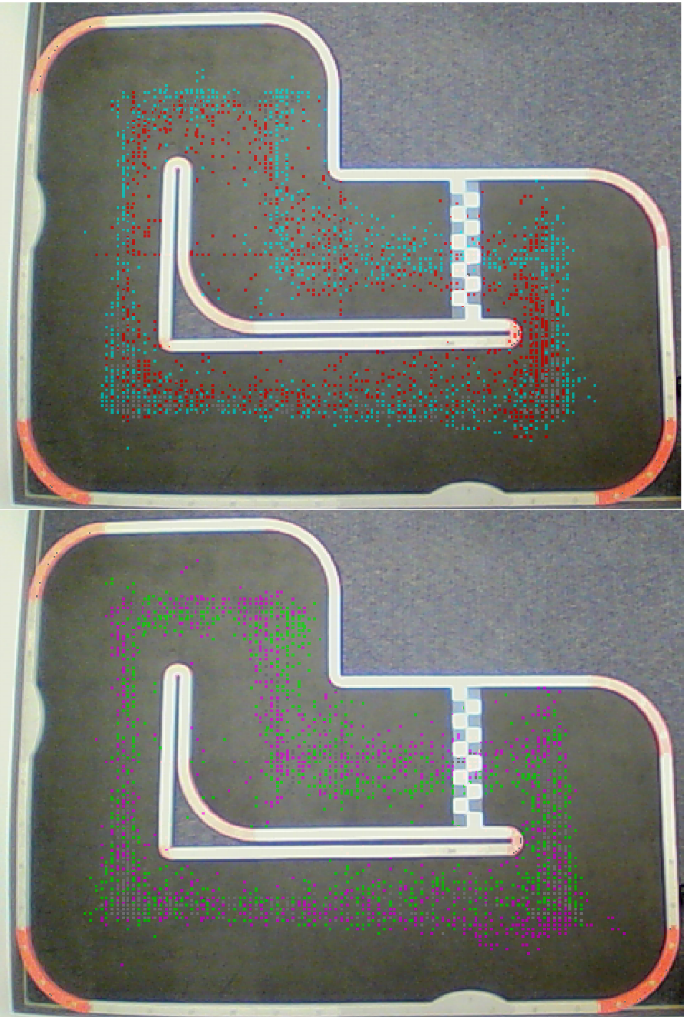}
\caption{Each dot represents the center of the vehicle at that point on the track at some point during the 40 test trials in testing phase one. Top) VGG-16 trained on framestack (cyan) and Inception-V3 trained on framestack (red). Bottom) AlexNet trained on single grayscale frame (purple) and Inception-V3 trained on single color frames (green).}
\label{fig:path_vis}
\end{centering}
\end{figure}

\begin{figure}
\begin{centering}
\includegraphics[width=\linewidth]{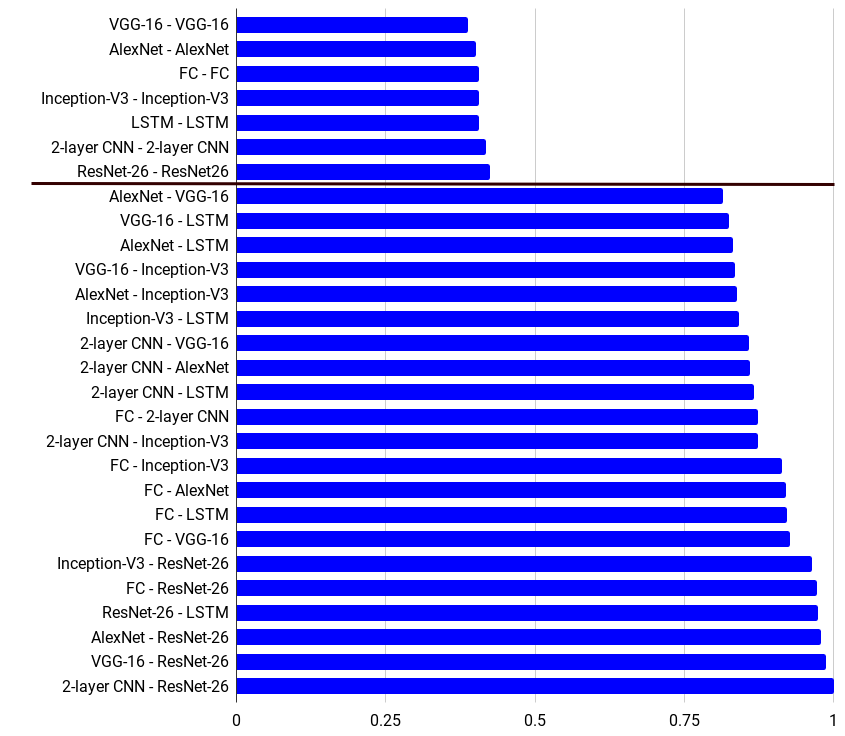}
\caption{The difference in path around the L-shaped track in testing phase one between each network over all image conditions.}
\label{fig:path_diff_bar}
\end{centering}
\end{figure}

In order to further compare driving performance across models, we used the video taken from the ceiling camera to track the specific paths taken by some of the networks highlighted in the section above over all test trials in testing phase one. These results also demonstrated the presence of a deployment gap in that similar training/validation did not always predict similar driving paths. For example, the path taken by Inception-V3 using grayscale framestack varied dramatically with that of VGG-16 using the same input, although they reached the same validation loss (Fig. \ref{fig:path_vis}, top). On the other hand, AlexNet, using single grayscale frames and Inception-V3 using single color frames obtained very different mean validation losses but the same success rate, and their paths were very similar (Fig. \ref{fig:path_vis}, bottom). A plot of pairwise similarity between all the networks (collapsed across all trials and image types) is shown in Figure \ref{fig:path_diff_bar}. The upper portion of the figure (above the horizontal black line) shows each model's similarity to itself across different loops around the track while the lower portion shows comparisons between different networks. As may be seen, each model's path exhibited much more self-similarity than similarity to the other models' paths. VGG-16 had the most self-similar (i.e. consistent) path, while ResNet-26 had the least self-similar path between trials. 

\begin{figure}
\begin{centering}
\includegraphics[width=\linewidth]{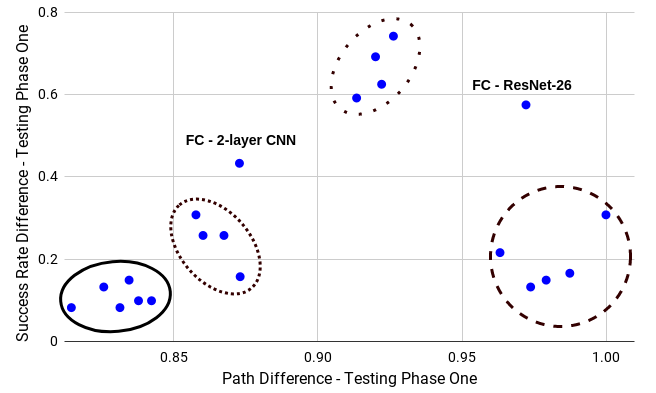}
\caption{Path difference across image types for all combinations of two networks. The solid ellipse in the bottom left contains all pairwise comparisons between AlexNet, VGG-16, Inception-V3, and LSTM. The ellipse to its right contains comparisons between the 2-layer CNN and AlexNet, VGG-16, Inception-V3, and AlexNet. The ellipse at the top contains comparisons between the fully-connected network and AlexNet, VGG-16, Inception-V3, and LSTM. The circle in the lower right contains comparisons between ResNet-26 and every other network except the fully-connected.}
\label{fig:path_diff_scatter}
\end{centering}
\end{figure}

Fig. \ref{fig:path_diff_scatter} shows the relationship between measures of path difference and success difference across all model pairs. As may be seen, there is a moderate positive trend such that models with similar paths had similar success rates. However, this trend does not characterize the data well. Instead, there appear to be four main `clusters', showing widely varying relationships, which we have outlined in the figure. The solid ellipse in the bottom left (models with similar paths and driving performance) contains all pairwise comparisons between AlexNet, VGG-16, Inception-V3, and LSTM. These were the highest-performing networks in both testing phases. The dashed ellipse to its right (moderately similar paths and success rates) contains comparisons between the 2-layer CNN on one hand and AlexNet, VGG-16 and Inception-V3 and LSTM on the other hand. The ellipse at the top (large differences in paths and success) contains comparisons between the fully-connected network on one hand and AlexNet, VGG-16, Inception-V3, and LSTM on the other. Finally, the circle in the lower right (very different paths but similar success rates) contains comparisons between ResNet-26 and every other network except the fully-connected.

\subsection{Inference Rate}
\begin{figure}
\begin{centering}
\includegraphics[width=\linewidth]{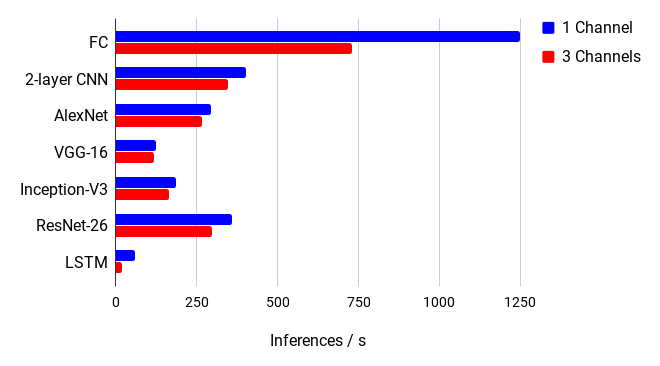}
\caption{The number of inferences each network was able to perform on color images.}
\label{fig:inf_rate}
\end{centering}
\end{figure}

The Fully-Connected network was able to perform the most inferences s\textsuperscript{-1} by far (729.83 inferences s\textsuperscript{-1} for color images; Fig. \ref{fig:inf_rate}), while the LSTM performed just 18 inferences s\textsuperscript{-1} on images with three channels. Generally, the larger, more advanced networks exhibited decreased inference rates than the smaller, more primitive ones (with the exception of ResNet-26 due to its relatively efficient architecture).

\subsection{Network ``Bias''}
\begin{figure}
\begin{centering}
\includegraphics[width=\linewidth]{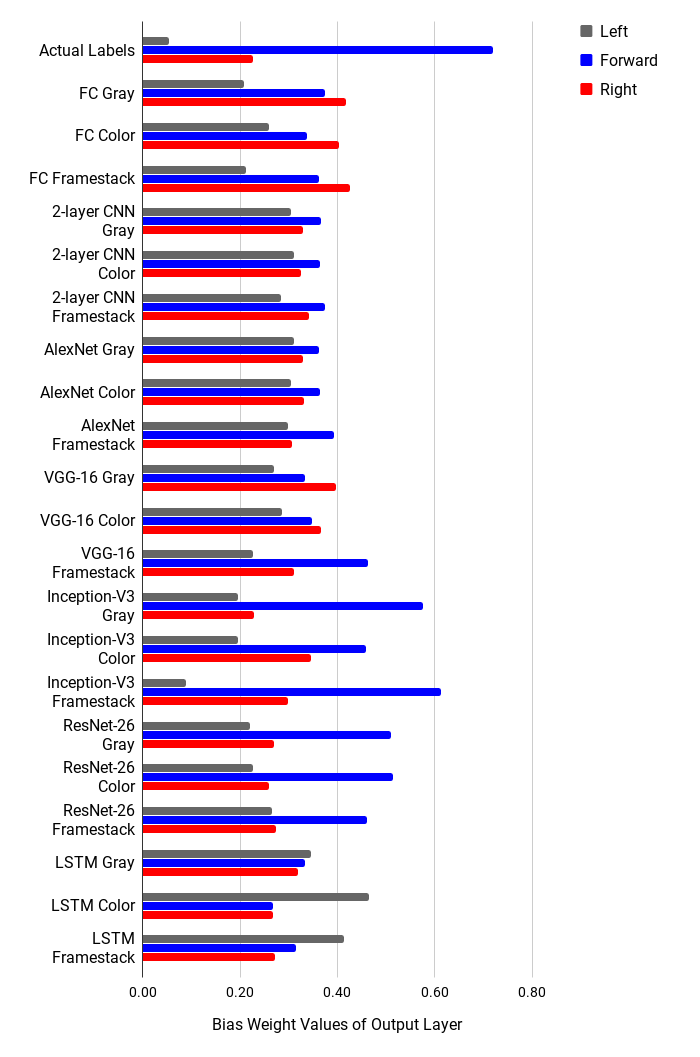}
\caption{The bias weights of each network's output layer, and the actual distribution of the dataset's labels (top). The backward action was not included due to its very low frequency relative to the other actions.}
\label{fig:bias_weights}
\end{centering}
\end{figure}

The bias weights of each network indicated that the networks that performed better were less 'biased' toward a particular action(s) (Fig. \ref{fig:bias_weights}, as the bias weights of their output layer were relatively similar. Most high-performing networks had dissimilar weight distributions relative to the labels, but some high-performing networks also had relatively similar weight distributions to the labels (e.g. Inception-V3 using single grayscale frame inputs).

\subsection{Hyper-Parameter Analysis}
\begin{figure}
\begin{centering}
\includegraphics[width=\linewidth]{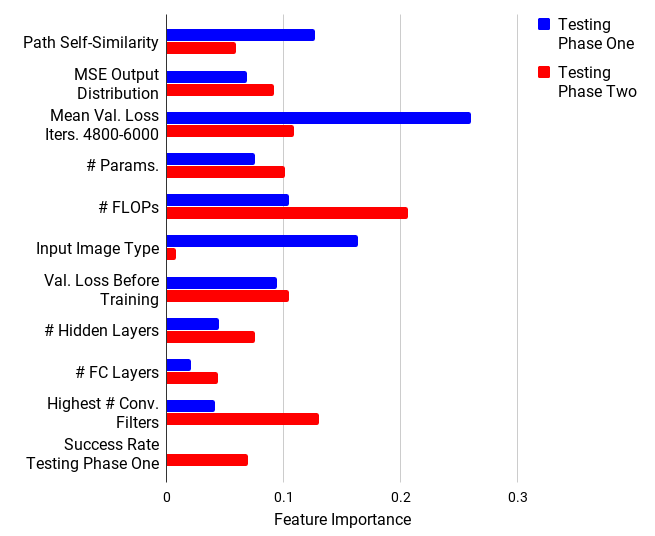}
\caption{The importance of each hyper-parameter in predicting the success rate in testing phases one (blue) and two (red) as determined by the random forest.}
\label{fig:feat_imp}
\end{centering}
\end{figure}

The levels of feature importance produced by the random forest analysis indicated that the validation loss was the most important feature in determining success in testing phase one, while the input image type was the second most important followed by path self-similarity and the number of FLOPs (Fig. \ref{fig:feat_imp}). For testing phase two, the number of FLOPs emerged as the most important feature, while the maximum number of convolutional filters in the network and validation loss were second and third most important respectively.

\subsection{Spatial Distribution of Attention}
\begin{figure}[ht]
\begin{centering}
\includegraphics[width=\linewidth]{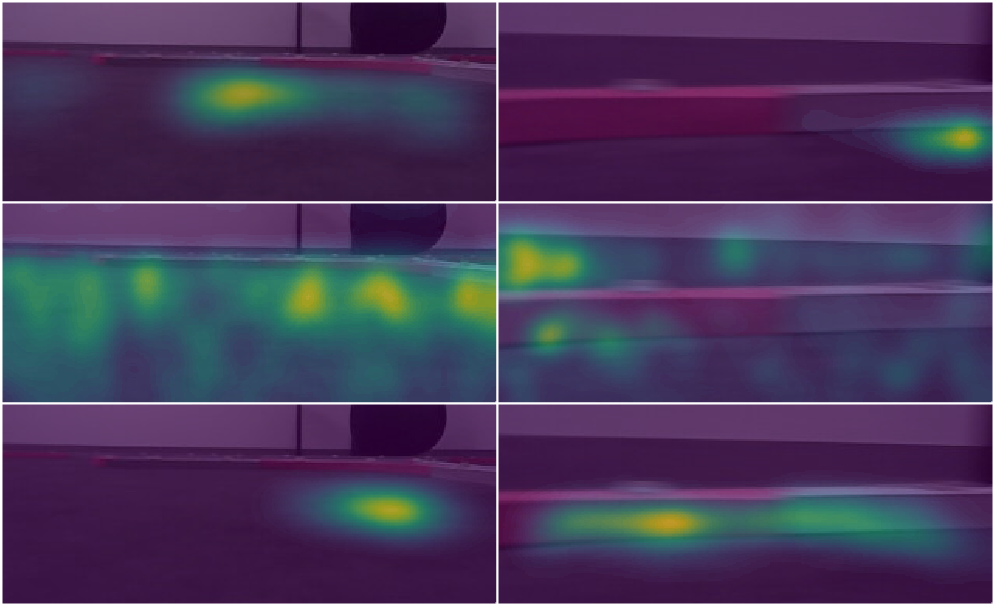}
\caption{Representative heatmaps depicting how much each pixel, when maximally flipped, changed the output of the network. The three heatmaps in the left column were created using the same image where the desired action was forward and the networks displayed are VGG-16 (top), LSTM (middle), and Inception-V3 (bottom). The three in the right column were all created from an image which had a desired action of right, and the networks displayed are AlexNet (top), 2-layer CNN (middle), and ResNet-26 (bottom). Each individual heatmap's values were scaled between 0 and 255, so intensities cannot be compared between heatmaps here.}
\label{heatmap_examples}
\end{centering}
\end{figure}

Table 1 illustrates the results of the pixel flipping analysis described above. Each metric contained within the table represents an average over 50 random test images. The action decisions of the two non-convolutional networks --- fully-connected and LSTM --- were completely unaffected by the flipped pixels (i.e. no pixel flip caused the network to change its action decision). Regarding the convolutional networks tested, the models that exhibited higher performance in both testing phases generally had more action decisions changed due to flipped pixels. Furthermore, the MSE between the output layers associated with the image containing the altered pixel and the unaltered image showed a similar trend, as the fully-connected and LSTM networks showed little difference. The convolutional networks, on average, contained much higher differences than the fully-connected-based networks. Despite having more decisions altered when given altered images, the convolutional networks were also more confident in their decisions in general when given unaltered images. Fig. \ref{heatmap_examples} depicts representative examples of some of the heatmaps constructed for single images using the MSE values for each pixel when flipped. Although VGG-16 (Fig. \ref{heatmap_examples}, top left) and AlexNet (Fig. \ref{heatmap_examples}, top right) had the most action decisions affected by flipped pixels per image on average, these pixels tended to lie in a very confined region of the image. This was not true for the LSTM (middle left) or the 2-layer CNN (middle right), as the information they attended to was fairly distributed, although the 2-layer CNN was affected more by pixels that did not correspond to `useful' features of the scene (e.g. objects outside of the track or parts of the room's wall such as the rubber baseboard).

\begin{table}[ht]
    \caption{Summary of Spatial Information Analysis}
    \centering
    \begin{tabular}[width=\linewidth]{c c c c}
    \hline \hline
    \\

    \textbf{Network} & \textbf{Num. Altered Actions} & \textbf{Output MSE} & \textbf{Confidence} \\
    \hline
    \\ FC & 0.00 & $4.08\times10$\textsuperscript{-6} & 0.80 \\
    2-layer CNN & 32.24 & $4.00\times10$\textsuperscript{-4} & 0.84 \\
    AlexNet & 881.92 & $1.30\times10$\textsuperscript{-3} & 0.91 \\
    VGG-16 & 880.88 & $2.00\times10$\textsuperscript{-3} & 0.92 \\
    Inception-V3 & 445.96 & $6.30\times10$\textsuperscript{-3} & 0.91 \\
    ResNet-26 & 270.46 & $1.00\times10$\textsuperscript{-3} & 0.85 \\
    LSTM & 0.00 & $9.87\times10$\textsuperscript{-6} & 0.86 \\
    \hline \hline
    \end{tabular}
\end{table}

\section{Discussion}

The current study presents the first systematic assessment and comparison of multiple  neural network models in an experimentally controlled driving task. Overall, we found that, with the exception of the fully-connected network, the most `primitive' among models, all networks achieved a reasonably good range of performance (80\% to 100\%) during testing phase one for both the color and grayscale input types and most performed very well ($\sim95$ \%) on at least one input type. The sole exception among contemporary networks was Resnet-26, which barely broke 80\% on any input type. With the exception of VGG-16, the inclusion of framestack data rather than single frames did not improve performance but actually reduced it, sometimes dramatically depending on the network. 

The presence of good performance across most contemporary models, at least for some data types, indicates that these models possess the computational complexity needed to perform the driving task. However, there was also a high degree of variability across models for specific data types. Critically, this variability was not well predicted by the models' validation performance (particularly in testing phase two), which was uniformly good across all models (except fully-connected) for all input types. These data demonstrate the presence of a large deployment gap.

Overall, the top performer among the tested group of models was AlexNet, using color images as an input, which outperformed every other network in both testing phase one and testing phase two. VGG-16 was the most robust to the input image type, as it achieved the same, high success rate across all three image types in testing phase one and had the second-best success rate in testing phase two. It also had a relatively small deployment gap, as its validation loss correlated relatively well with its success rate in both testing phases. Inception-V3 using color images as input also had a relatively small deployment gap, as it had the lowest validation loss out of all networks and its success rates in both testing phases were comparably good. The single, color image yielded the best success rates when used as input to all networks except the LSTM. The single, grayscale images consistently performed slightly worse, but still well, on average compared to the single color frames. The gray framestack, however, yielded the poorest and most inconsistent results out of the three, although many of the contemporary networks obtained decent success rates using it. 

The path analysis demonstrated that the route taken over the testing trials was much more similar within models than between models and the better performing networks generally took a more consistent path across trials than the worse performing ones. Furthermore, these more successful models tended to converge on a relatively similar path around the track. In contrast, the fully-connected network and ResNet-26 both took very different paths than the other networks. In the case of the fully-connected, this different path led to very poor performance while in the case of ResNet-26, it led to moderately worse performance. This suggests that both models very quickly entered new `terrain' never encountered before by the human driver, but the latter was able to generalize more (mainly in test phase one) while the former was not. The performance of ResNet-26 is explored in greater detail below.  

The results of the pixel-flipping analysis illustrate that the networks that do not use convolutional layers, such as the fully-connected network and LSTM, are affected far less by flipped pixels than networks that do use convolutions. However, they also do not perform as well on the task. One explanation for this is that the fully-connected networks by nature look for global patterns in the input, whereas CNNs look for local patterns via convolutions, and, thus, a local perturbation in the input should not have as strong an effect on the performance of the former, as we report (Table 1). Knowing what specifically to look at in the image allows CNNs to be more efficient/better at image recognition tasks than networks without convolutions, but as a result, this also makes these networks less robust to certain types of noise (e.g. adversarial attacks).

One of the most interesting and important results was evidence for the presence of a significant deployment gap --- that is, models that performed similarly well during validation showed highly variable performance during testing. At first glance, this may seem puzzling. If a model can generalize from training to validation then why not from training to the images during deployment?  We believe this deployment gap may be explained by the fact that each inference in a self-driving task is not causally isolated from future inferences, as in a traditional image recognition task. Instead, each inference made by the network leads to a behavioral choice, which, in turn, affects  \textit{all subsequent inputs}. This means that a small initial difference in the path the vehicle takes can lead it on a novel `orbit' unlike any paths taken by a human driver. In turn, this means the resulting \textit{inputs} will be  different than those present in the training or validation set (all of which were human-driver generated) and errors in generalization may accumulate, leading to a vicious circle of unfamiliar inputs and resulting actions.

One factor that could effect a networks performance in response to novel inputs is the extent to which it has a bias to choose any particular action, which can be due to that action's frequency in the training set. If a network does have a strong bias, it is likely to choose this action most frequently. This can be evidenced in the `bias' analysis of each network as shown in Fig. \ref{fig:bias_weights}, where the bias weight of each output node was examined for every network. Although a few exceptions were present, the networks that performed better in both testing phases (e.g. AlexNet, VGG-16, LSTM) were less `biased' toward a particular action, which indicates that they were more easily able to adapt to a situation in which the action distribution was very different from the training set due to a previously untraveled orbit. In this sense, the preference of one action much more than the others as indicated by the metrics used in the bias analysis illustrate a form of overfitting, as the bias weights of the output layer are relied upon too heavily (possibly to the detriment of the rest of the weights in the network). Furthermore, these same networks, with the addition of Inception-V3, were the best-performing networks in both testing phases, and they all converged on a comparably similar path (Fig. \ref{fig:path_diff_scatter}). Therefore, they had to generalize less because they took a more consistent path, which was also taken by other good networks (Figures \ref{fig:path_diff_bar} and \ref{fig:path_diff_scatter}), but they were also able to generalize more because they were not biased too much toward a particular action.  

One surprising, and somewhat disappointing, result was the poor performance of most models on the gray framestack in testing phase one compared with the other two input types (Fig. \ref{fig:success_rates}, top), and in every case --- except for VGG-16 where it accrued an equivalent success rate as the other input types --- it performs worse than a single gray frame. This seems counter-intuitive because more information, in the form of past images of the same type, would be expected to allow a network to achieve equivalent or even better performance, than single frames. This poor performance cannot be attributed to the increased size of the input images from one channel to three, because in most cases, a single color frame --- which also has three channels --- had the best success. 

%This procedure was performed five times for both color and framestack image types, and the results were averaged.

We believe that this poor performance may be due to the fact that a stack of three frames may carry more complexity than a single frame or three color channels, leading to greater computational strain on the model. To test this, we calculated the Structural Similarity Index (SSIM) \cite{wang2004image} --- which is commonly used as a measure of image similarity --- between each channel in the color images and each channel in the framestack images respectively, averaged the three, then took the average over 1000 random images. The average SSIM between color image channels over these five runs was 0.94, and the average SSIM between framestack channels was 0.68. Therefore, it is possible to conclude that the gray framestack images performed poorly relative to the other image types, and extremely poorly in some of the simpler networks, because the channels in this type of image are not as correlated as those in color images, adding complexity. We believe this is also made worse by the fact that it is impossible to drive exactly the same as in the training dataset (as detailed in the paragraph above), meaning the past frames will never be exactly the same as those the network was trained on, and the network will need to generalize even further to overcome this. This is likely why most of the more recent, contemporary networks perform adequately when using framestack images, but the fully-connected network and 2-layer CNN perform very badly.

The performance of the ResNet-26 architecture may exemplify the challenge of the deployment gap. This architecture has been used widely in many image-related networks, from classifiers to generative networks, mainly due to its ability to attain very good classification accuracies on many computer vision benchmarks while using less weights and complexity than many other contemporary networks, such as VGG and Inception. In fact, prior research comparing some of these same networks on object detection found that ResNet-50 outperformed both Inception-V3 and AlexNet \cite{comparison_nn_2018}, both of which outperformed ResNet in this experiment. One possible explanation is that, while relatively deep compared to the other contemporary networks tested, our ResNet-26 architecture here is not as wide as most, as the maximum number of filters it has in a single layer is 128. For example, the 2-layer CNN --- which outperformed ResNet-26 during testing phase one when using the gray and color single frames and was not greatly outperformed by ResNet in testing phase two --- has twice the number of filters as ResNet in its second hidden layer (256) but is much more shallow. It seems reasonable to conclude that these additional hidden layers relative to most other networks tested here should have been able make up for this deficit in width, as it has been reported that depth is more important than width in determining representational power and ability of a network to approximate a given function \cite{eldan2016power}\cite{liang2017deep}\cite{safran2016depth}. However, considering that the number of FLOPs in the network was found to play a relatively large role in success during both testing phases, and the highest number of convolutional filters in a CNN's hidden layer was especially significant in testing phase two, it is likely that these two probably had some role in ResNet-26's performance.

Instead, we believe this network's relatively poor performance is likely due to its unusual orbit (its paths show low self-similarity as well as low similarity to the high-performing models, as shown in Fig. \ref{fig:path_diff_bar}) coupled with its strong bias as shown in Figure \ref{fig:bias_weights}.  This means that this network was not very consistent, which would cause its inputs and outputs to be different than those in the training set as it traversed the track. This would require it to generalize more in order to perform well; however its bias toward certain actions, based on training, precludes successful adaptation to this new, previously unseen path.

%For insight into this question, we turn to the work of \cite{DBLP:journals/corr/abs-1709-06247}, which re-examined the 1:1 convolution to relu paradigm of CNNs. 

% However, it is deeper than most of the other networks compared in this research, which should have helped make up for the low width of the network.

Another interesting result was the LSTM network's superior performance on the grayscale inputs compared to the color frames, something not seen in other networks. This may be explained quite simply with the aid of Fig. \ref{fig:inf_rate}. On images the same size and type as those coming through the vehicle's camera, the LSTM used here was only able to perform 18 inferences s\textsuperscript{-1} on images with three channels, such as the single color frame and the gray framestack, but for the single gray image, it was able to perform 60 inferences s\textsuperscript{-1}. Since the vehicle operated at 30 FPS during the course of this research, it was impossible for the LSTM using color and framestack inputs to produce inferences fast enough, which probably caused dropped frames and/or bottlenecking.

% SHOULD WE LEAVE THIS IN?????
%##########################################################################
%One surprising finding of this research is the relative importance of the starting validation loss to the network's success (Fig. 5). For insight into possible causes of this, we turn to \cite{saxe2011random} and \cite{sapiro}. The first study concludes that random weights help evaluate candidate architectures for a given task, as better architectures often perform better using random weights, while the second study illustrates the ubiquity of random Gaussian weights in classification tasks using CNNs. Given these two results, it is possible that AlexNet happens to be the architecture best suited for this particular task and setting, as it had a much lower validation loss than the others before any training occurred (Fig. 5).
%##########################################################################

\section{CONCLUSION}
In summary, the present study presents the first systematic comparison of multiple contemporary deep learning architectures and training protocols in a self-driving vehicle. While most contemporary architectures showed reasonably good driving performance for some input types, we found that there was often a gap between performance on the validation dataset and actual driving, which we term the `deployment gap'. We believe this gap is likely based on the fact that small initial differences in driving behavior can lead to inputs that are poorly represented by the human-generated training/validation data, which requires further generalization. This is likely why most of the networks that performed the best, such as AlexNet, VGG-16, and Inception-V3, seemed to take similar paths around the track and had relatively consistent paths over multiple runs. Conversely, the ResNet-26 architecture did not take a very consistent path around the track, its path was very different than all other networks, and it was also heavily biased toward certain actions, which decreased its ability to generalize to these new inputs. Based on its validation performance alone --- which was above average and similar to other networks that performed well in the testing phases --- there was no indication that its performance would be so different when deployed in an embedded control system. Thus, this network demonstrates that the deployment gap exists and needs to be taken into consideration when designing real world end-to-end DL-based systems. In addition to demonstrating a deployment gap, this research also suggests some potential ways to identify and address such gaps including an even bias across output actions, increased FLOPs and convolutional filters, and an increased weighting of useful features in the input images. These may be more useful than validation performance in determining driving ability when a deep neural network is used to control a vehicle in an end-to-end fashion. 

%%%%%%%%%%%%%%%%%%%%%%%%%%%%%%%%%%%%%%%%%%%%%%%%%%%%%%%%%%%%%%%%%%%%%%%%%%%%%%%%
\section{Acknowledgments}
The authors gratefully acknowledge the support of NVIDIA Corporation with the donation of the GPU hardware used for this research. The authors would also like to thank Levi Stein for his hard work and creativity throughout this research.

\bibliographystyle{unsrt}
\bibliography{reference}

% \begin{table}
%   \vspace*{7in}
%   \caption{Layer architecture of the included models}
%   \label{tbl:excel-table}
%   \includegraphics[width=\linewidth]{network_table.png}
% \end{table}

\vspace*{1in}
\begin{landscape}
\begin{table}
\begin{centering}
\caption{A detailed description of the architecture of each network used in this research and its number of FLOPs and parameters.}
\includegraphics[width=21.0cm]{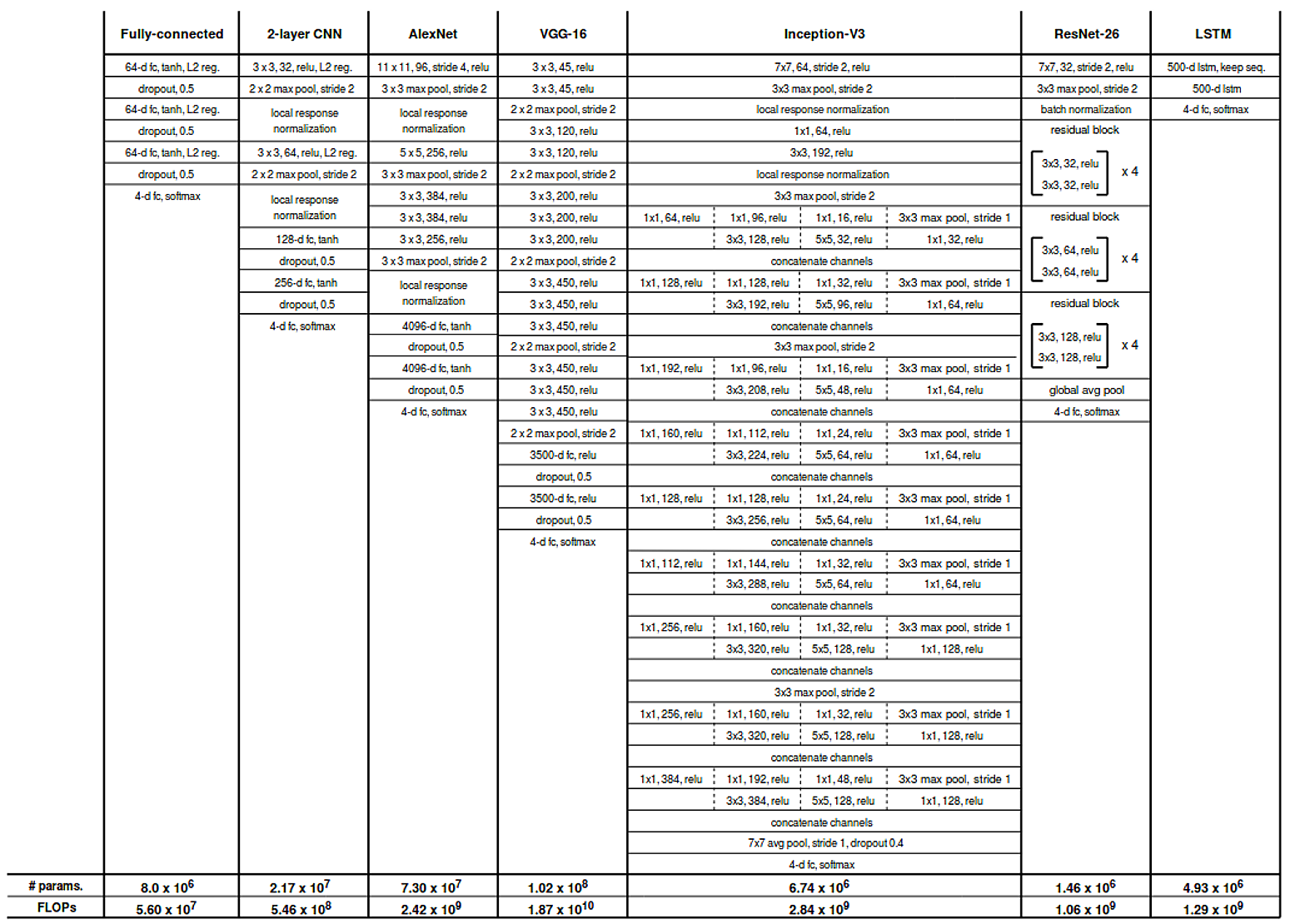}
\begin{center}
\end{center}
\end{centering}
\label{fig:network_graph}
\end{table}
\end{landscape} %
%%%%%%%%%%%%%%%%%%%%%%%%%%%%%%%%%%%%%%%%%%%%%%%%%%%%%%%%%%%%%%%%%%%%%%%%%%%%%%%

\end{document}